\documentclass[12pt]{article}

\usepackage[auth-sc,affil-sl]{authblk}
\usepackage{graphicx}
\usepackage{color}
\usepackage{amsmath}
\usepackage{dsfont}
\usepackage[round]{natbib}
\usepackage{placeins}
\usepackage{amssymb}
\usepackage{abstract}
\usepackage{hyperref}
\usepackage{cancel}
\usepackage[margin=1.05in]{geometry}
\usepackage{enumerate}
\usepackage{listings}
\usepackage{mathdots}
\usepackage{array}
\usepackage{phaistos}
\usepackage{textcomp}
\usepackage{subcaption}
\usepackage{algorithm}
\usepackage{algorithmicx}
\usepackage{algpseudocode}

\newcommand{\qu}[1]{``#1''}

\newcommand{\treet}[1]{\text{\scriptsize \PHplaneTree}_{#1}}
\newcommand{\treeleaft}[1]{\text{\scriptsize \PHplaneTree}_{#1}^{\text{\tiny \textleaf}}}
\newcommand{\leaf}{\text{\scriptsize \textleaf}}

\newcounter{probnum}
\allowdisplaybreaks

\definecolor{gray}{rgb}{0.7,0.7,0.7}

\definecolor{black}{rgb}{0,0,0}
\definecolor{white}{rgb}{1,1,1}
\definecolor{blue}{rgb}{0,0,0.7}
\newcommand{\inblue}[1]{\color{blue}\textbf{#1} \color{black}}
\definecolor{green}{rgb}{0.133,0.545,0.133}
\newcommand{\ingreen}[1]{\color{green}\textbf{#1} \color{black}}
\definecolor{yellow}{rgb}{1,0.549,0}
\newcommand{\inyellow}[1]{\color{yellow}\textbf{#1} \color{black}}
\definecolor{red}{rgb}{1,0.133,0.133}
\newcommand{\inred}[1]{\color{red}\textbf{#1} \color{black}}
\definecolor{purple}{rgb}{0.58,0,0.827}

\definecolor{brown}{rgb}{0.55,0.27,0.07}

\definecolor{backgcode}{rgb}{0.97,0.97,0.8}
\definecolor{Brown}{cmyk}{0,0.81,1,0.60}
\definecolor{OliveGreen}{cmyk}{0.64,0,0.95,0.40}
\definecolor{CadetBlue}{cmyk}{0.62,0.57,0.23,0}

\newcommand{\bv}[1]{\boldsymbol{#1}}

\newcommand{\sigsq}{\sigma^2}

\newcommand{\thetavec}{\bv{\theta}}

\newcommand{\iid}{~{\buildrel iid \over \sim}~}

\newcommand{\B}{\bv{B}}

\newcommand{\M}{\bv{M}}

\newcommand{\X}{\bv{X}}

\newcommand{\I}{\bv{I}}

\newcommand{\Y}{\bv{Y}}

\newcommand{\x}{\bv{x}}

\newcommand{\zerovec}{\bv{0}}

\newcommand{\y}{\bv{y}}

\newcommand{\threebythreemat}[9]{\bracks{\begin{array}{ccc} #1 & #2 & #3 \\ #4 & #5 & #6 \\ #7 & #8 & #9 \end{array}}}

\newcommand{\beqn}{\vspace{-0.25cm}\begin{eqnarray*}}
\newcommand{\eeqn}{\end{eqnarray*}}
\newcommand{\bneqn}{\vspace{-0.25cm}\begin{eqnarray}}
\newcommand{\eneqn}{\end{eqnarray}}

\newcommand{\parens}[1]{\left(#1\right)}

\newcommand{\prob}[1]{\mathbb{P}\parens{#1}}
\newcommand{\cprob}[2]{\prob{#1~|~#2}}

\newcommand{\bracks}[1]{\left[#1\right]}
\newcommand{\braces}[1]{\left\{#1\right\}}

\newcommand{\expe}[1]{\mathbb{E}\bracks{#1}}
\newcommand{\cexpe}[2]{\expe{#1 ~ | ~ #2}}

\newcommand{\corr}[2]{\text{Corr}\bracks{#1, #2}}

\newcommand{\multnormnot}[3]{\mathcal{N}_{#1}\parens{#2,\,#3}}
\newcommand{\normnot}[2]{\mathcal{N}\parens{#1,\,#2}}

\newcommand{\errorrv}{\mathcal{E}}
\newcommand{\berrorrv}{\bv{\errorrv}}

\newcommand{\Xobs}{{\X_{\text{obs}}}}

\newcommand{\Xminjmiss}{\X_{-j,\text{miss}}}
\newcommand{\Xminjobs}{\X_{-j,\text{obs}}}

\newcommand{\gammavec}{\bv{\gamma}}

\newcommand{\Xtrain}{\X_{\text{train}}}
\newcommand{\ytrain}{\y_{\text{train}}}

\title{Prediction with Missing Data via Bayesian Additive Regression Trees}
\author{Adam Kapelner\footnote{Both authors contributed equally to this work.}~\thanks{Electronic address: \texttt{kapelner@wharton.upenn.edu}; Prinicipal Corresponding author}~}
\author{Justin Bleich$^*$\thanks{Electronic address: \texttt{jbleich@wharton.upenn.edu}; Corresponding author}}
\affil{The Wharton School of the University of Pennsylvania}

\begin{document}
\maketitle

\begin{abstract}
We present a method for incorporating missing data into general forecasting problems which use non-parametric statistical learning. We focus on a tree-based method, Bayesian Additive Regression Trees (\texttt{BART}), enhanced with \qu{Missingness Incorporated in Attributes,} an approach recently proposed for incorporating missingness into decision trees. This procedure extends the native partitioning mechanisms found in tree-based models and does not require imputation. Simulations on generated models and real data indicate that our procedure offers promise for both selection model and pattern mixture frameworks as measured by out-of-sample predictive accuracy. We also illustrate \texttt{BART}'s abilities to incorporate missingness into uncertainty intervals. Our implementation is readily available in the \texttt{R} package \texttt{bartMachine}.
\end{abstract}

\section{Introduction}

This article addresses prediction problems where covariate information is missing during model construction and is also missing in future observations for which we are obligated to generate a forecast. Our aim is to innovate a non-parametric statistical learning extension which incorporates missingness into \textit{both} the training and the forecasting phases. In the spirit of non-parametric learning, we wish to incorporate the missingness in both phases automatically, without the need for pre-specified modeling. 

We limit our focus to tree-based statistical learning, which has demonstrated strong predictive performance and has consequently received considerable attention in recent years. State-of-the-art algorithms include Random Forests \citep[\texttt{RF},][]{Breiman2001a}, stochastic gradient boosting \citep{Friedman2002}, and Bayesian Additive and Regression Trees \citep[\texttt{BART},][]{Chipman2010}, the algorithm of interest in this study. Popular implementations of these methods do not incorporate covariate missingness \textit{natively} without relying on either imputation or a complete case analysis of observations with no missing information.

Previous simulations and real data set applications have indicated that \texttt{BART} is capable of achieving excellent predictive accuracy. Unlike most competing techniques, \texttt{BART} is composed of a probability model, rather than a procedure that is purely ``algorithmic'' \citep{Breiman2001}. \texttt{BART} presents an alternative approach to statistical learning for those comfortable with the Bayesian framework. This framework provides certain advantages, such as built-in estimates of uncertainty in the form of credible intervals as well as the ability to incorporate prior information on covariates \citep{Bleich2013}. However, no means for incorporating missing data in \texttt{BART} has been published to date. Our goal here is to develop a principled way of adapting \texttt{BART}'s machinery to incorporate missing data that takes advantage of the Bayesian framework.

Our proposed method, henceforth named \texttt{BARTm}, modifies the recursive partitioning scheme during construction of the decision trees to incorporate missing data into splitting rules. By relying on the Metropolis-Hastings algorithm embedded in \texttt{BART}, our method attempts to send missing data to whichever of the two daughter nodes increases overall model likelihood. Missingness \textit{itself} also becomes a valid splitting criterion. 

During model construction, taking advantage of this modified set of splitting rules does not require imputation, which relies on assumptions that cannot be easily verified. Our approach is equally viable for continuous and nominal covariate data and both selection and pattern mixture models. The latter models do not assume that missing data is necessarily free of information; the data may have gone missing for a reason crucial to the response function and therefore crucial to our forecast. \texttt{BARTm} is able to exploit this relationship when appropriate.

Since missingness is handled natively within the algorithm, \texttt{BARTm} can generate predictions on future data with missing entries as well. Additionally, \texttt{BART}'s Bayesian framework also naturally provides estimates of uncertainty in the form of credible intervals. The amount of uncertainty increases with the amount of information lost due to missingness; thereby missingness is appropriately incorporated into the standard error of the prediction. Also, our proposed procedure has negligible impact on the runtime during both model construction and prediction phases. 

In Sections \ref{subsec:overview} - \ref{subsec:decision_tree_missing}, we provide a framework for statistical learning with missingness with a focus on decision trees. We give a brief overview of the \texttt{BART} algorithm in Section \ref{subsec:bart} and explain the internals of \texttt{BARTm} in Section \ref{sec:mia_in_bart}. We then demonstrate \texttt{BARTm}'s predictive performance on generated models in Section \ref{sec:generated_data_sims} as well as real data with a variety of missing data scenarios in Section \ref{sec:real_data_sims}. We conclude with Section \ref{sec:discussion}. \texttt{BARTm} can be found in the \texttt{R} package \texttt{bartMachine} which is available on CRAN \citep{Kapelner2013}.

\section{Background}\label{sec:background}

\subsection{A Framework for Missing Data in Statistical Learning}\label{subsec:overview}

Consider $p$ covariates $\X := \bracks{X_1, \ldots, X_p}$, a continuous response $\Y$\footnote{When $\Y$ is binary, we use a similar framework with an appropriate link function encapsulating $f$.} and an unknown function $f$ where $\Y = f(\X) + \berrorrv$. We denote $\berrorrv$ as the noise in the response unexplained by $f$. The goal of statistical learning is to use the \textit{training set}, $\bracks{\ytrain, \Xtrain}$ which consists of $n$ observations drawn from the population $\prob{\Y,~\X}$, to produce an estimate, $\hat{f}$, the best guess of $\cexpe{\Y}{\X}$. This function estimate can then be used to generate predictions on future test observations with an unknown response. We denote these future observations as $\X_*$ which we assume are likewise drawn from the same population as the training set.

Missingness is one of the scourges of data analysis, plaguing statistical learning by causing missing entries in both the training matrix $\Xtrain$ as well as missing entries in the future records, $\X_*$. In the statistical learning context, the training set is defined by observations which do not exhibit missingness in their response, $\ytrain$. Records with missing responses cannot be used to build models for estimation of $f$.\footnote{\qu{Imputing missing values in the response} for the new $\X_*$ is equivalent to \qu{prediction} and is the primary goal of statistical learning.} Thus, \qu{missingness} considered in this paper is missingness \textit{only} in $\Xtrain$ and $\X_*$. We denote missingness in the $p_M \leq p$ features of $\X$ which suffer from missingness as $\M := \bracks{M_1, \ldots, M_{p_M}}$, binary vectors where 1 indicates missing and 0 indicates present, and covariates that are present with $\Xobs := \bracks{X_{\text{obs}_1}, \ldots, X_{\text{obs}_p}}$. The main goal of statistical learning with missingness is to estimate $\cexpe{\Y}{\Xobs, \M}$.

We now frame missing data models in statistical learning using the canonical framework of selection and pattern-mixture models \citep{Little1993}. Conditional on $\X$, \textit{selection models} factor the full data likelihood as

\bneqn\label{eq:selection_model}
\cprob{\Y, \M}{\X, \thetavec, \gammavec} = \cprob{\Y}{\X, \thetavec} \cprob{\M}{\X, \gammavec}
\eneqn

\noindent where $\thetavec$ and $\gammavec$ are parameter vectors and are assumed distinct. The first term on the right hand side reflects that the marginal likelihood for the response $\cprob{\Y}{\X, \thetavec}$ is independent of missingness. The second term on the right conventionally conditions on $\Y$. In the forecasting paradigm, missingness is \textit{assumed independent} of the response because $\Y$ is often yet to be realized and thus its unknown value should not influence $\M$, the missingness of the previously realized covariates.

Conditional on $\X$, \textit{pattern mixture models} partition the full data likelihood as 

\bneqn\label{eq:pattern_mixture_model}
\cprob{\Y, \M}{\X, \thetavec, \gammavec} = \cprob{\Y}{\M, \X, \thetavec} \cprob{\M}{\X, \gammavec}.
\eneqn

\noindent where $\thetavec$ and $\gammavec$ are parameter vectors and again assumed distinct. The difference between the above and Equation~\ref{eq:selection_model} is the marginal likelihood of the response is now a function of $\M$. This means there can be different response models under different patterns of missingness in the $p_M$ covariates.

In both selection and pattern-mixture paradigms, the term on the right is the \textit{missing data mechanism} (MDM), which traditionally is the mechanism controlling missingness in the response. In our framework however, the MDM controls missingness only in $\X$: the covariates (and parameters $\gammavec$) create missingness within themselves which inevitably needs to be incorporated during model construction and forecasting. Thus, the MDM is conceptually equivalent in both the selection and pattern mixture paradigms.

The conceptual difference between the selection and pattern mixture models in the statistical learning framework can be envisioned as follows. Imagine the full covariates $\X$ are realized but due to the MDM, $\X$ is latent and we instead observe $\Xobs$ and $\M$. In the selection paradigm, $\Y$ is realized only from the full covariates via $\cprob{\Y}{\X, \thetavec}$. However, in the pattern-mixture paradigm, both $\X$ and $\M$ intermix to create many collated response models $\braces{\cprob{\Y}{\X, \thetavec, \M = m}}_{m \in \mathcal{M}}$ corresponding to different points in $\M$-space. Thus, under our assumptions, selection models are a subset of pattern mixture models. Note that pattern-mixture models are chronically under-identified and difficult to explicitly model in practice. We address why our proposed method is well-suited to handle prediction problems under this framework in Section~\ref{sec:mia_in_bart}.

We now present \citet{Little2002}'s taxonomy of MDM's which are traditionally framed in the selection model paradigm but here apply to both paradigms: (1) missing completely at random (MCAR), (2) missing at random (MAR) and (3) not missing at random (NMAR). MCAR is a mechanism that generates missingness in covariate $j$ without regard to the value of $X_j$ itself nor the values and missingness of any other covariates, denoted $\X_{-j}$. MCAR is exclusively determined by exogenous parameter(s) $\gammavec$. The MAR mechanism generates missingness without regard to $X_j$, its own value, but can depend on values of other attributes $\X_{-j}$ as well as $\gammavec$. The NMAR mechanism features the additional dependence on the value of $X_j$ itself as well as unobserved covariates.\footnote{Explicit dependence on unobserved covariates was not explored as MDM's in this paper.} We summarize these mechanisms in Table \ref{tab:mdms}. In our framework, each of the $p_M \leq p$ covariates with missingness, denoted as $X_j$'s, are assumed to have their own MDM$_j$. Thus, the full MDM for the whole covariate space, $\cprob{\M}{\X, \gammavec}$, can be arbitrarily convoluted, exhibiting combinations of MCAR, MAR and NMAR among its $p_M$ covariates and each MDM$_j$ relationship may be highly non-linear with complicated interactions.

\begin{table}[htp]
\centering
\begin{tabular}{l|l}
MDM & $\cprob{\M_j}{X_{j,\text{miss}}, \Xminjmiss, \Xminjobs, \gammavec} = \ldots$ \\ \hline
MCAR & $\cprob{\M_j}{\gammavec}$ \\ 
MAR & $\cprob{\M_j}{\Xminjmiss, \Xminjobs, \gammavec}$ \\ 
NMAR & (does not simplify)
\end{tabular}
\caption{MDM models in the context of statistical learning. $\M_j$ is an indicator vector which takes the value one when the $j^{\text{th}}$ covariate is missing for the $i$th observation. $\Xminjobs$ are the observed values of the other covariates, besides $j$. $\Xminjmiss$ are the values of the other covariates, besides $j$, which are not observed because they are missing.}
\label{tab:mdms}
\end{table}

We conclude this section by emphasizing that in the non-parametric statistical learning framework where predictive performance is the objective, there is no need for explicit inference of $\thetavec$ (which may have unknown structure and arbitrary, possibly infinite, dimension). Instead, the algorithm performs \qu{black-box} estimation of the data generating process such that the output $\hat{f}$ estimates the $\cexpe{\Y}{\Xobs, \M}$ function. Thus, if we can successfully estimate this conditional expectation function directly, then accurate forecasts can be obtained. This is the approach that \texttt{BARTm} takes.

\subsection{Strategies for Incorporating Missing Data}\label{subsec:missing_data_strategies}

A simple strategy for incorporating missingness into model building is to simply ignore the observations in $\Xtrain$ that contain at least one missing measurement. This is called \qu{list-wise deletion} or \qu{complete case analysis.} It is well known that complete case analysis will be unbiased for MCAR and MAR selection models where missingness does not depend on the response when the target of estimation is $\cexpe{\Y}{\X}$. However, when forecasting, the data analyst must additionally be guaranteed that $\X_*$ has no missing observations, since it is not possible to generate forecasts for these cases.

By properly modeling missingness, incomplete cases can be used and more information about $\cexpe{\Y}{\X}$ becomes available, potentially yielding higher predictive performance. One popular strategy is to guess or \qu{impute} the missing entries. These guesses are then used to \qu{fill in} the holes in $\Xtrain$ and $\X_*$. The imputed $\Xtrain$ is then used \textit{as if} it were the real covariate data when constructing $\hat{f}$ and the imputed $\X_*$ is then used as if it were the real covariate data during forecasting. To carry out imputation, the recommended strategy is to properly model the predictive distribution and then draws from the model are used to fill in the missing entries. \textit{Multiple imputation} involves imputing many times and averaging over the results from each imputation \citep{Rubin1978}. In statistical learning, a prediction could be calculated by averaging the predictions from many $\hat{f}$'s built from many imputed $\Xtrain$'s and then further averaging over many imputed $\X_*$'s. In practice, having knowledge of both the missing data mechanism and each probability model is very difficult and has usually given way to nonparametric methods such as $k$-nearest neighbors \citep{Troyanskaya2001} for continuous covariates and saturated multinomial modeling \citep{Schafer1997} for categorical covariates. The widely used \texttt{R} package \texttt{randomForest} \citep{randomForestPackage} imputes via ``hot-decking'' \citep{Little2002}.


A more recent approach, \texttt{MissForest} \citep{Stekhoven2012}, fits nonparametric imputation models for any combination of continuous and categorical input data, even when the response is unobserved. In this unsupervised procedure (i.e., no response variable needed), initial guesses for the imputed values are made. Then, for each attribute with missingness, the observed values of that attribute are treated as the response and a \texttt{RF} model is fit using the remaining attributes as predictors. Predictions for the missing values are made via the trained \texttt{RF} and serve as updated imputations. The process proceeds iteratively through each attribute with missingness and then repeats until a stopping criterion is achieved. The authors argue that their procedure intrinsically constitutes multiple imputation due to Random Forest's averaging over many unpruned decision trees. The authors also state that their method will perform particularly well when \qu{the data include complex interactions or non-linear relations between variables of unequal scales and different type.} Although no explicit reference is given to \citet{Little2002}'s taxonomy in their work, we expect \texttt{MissForest} to perform well in situations generally well-suited for imputation, namely, the MCAR and MAR selection models discussed in Section~\ref{subsec:overview}. \texttt{MissForest} would not be suited for NMAR MDMs as imputation values for $X_j$ can only be modeled from $\X_{-j}$ in their implementation. Additionally, implementing \texttt{MissForest} would not be recommended for pattern-mixture scenarios because imputation is insufficient to capture differing response patterns.

Since \texttt{BART} is composed primarily of a sum-of-regression-trees model, we now review strategies for incorporating missing data in tree-based models. 

\subsection{Missing data in Binary Decision Trees} \label{subsec:decision_tree_missing}

\textit{Binary decision trees} are composed of a set of connecting nodes. \textit{Internal nodes} contain a \textit{splitting rule}, for example, \texttt{x$_j$ < c}, where \texttt{x$_j$} is the \textit{splitting attribute} and \texttt{c} is the \textit{splitting value}. An observation that satisfies the rule is passed to the left daughter node otherwise it is passed to the right daughter node. This partitioning proceeds until an observation reaches a \textit{terminal node}. Terminal nodes (also known as \textit{leaves}) do not have splitting rules and instead have \textit{leaf values}. When an observation \qu{lands} in a terminal node it is assigned the leaf value of the terminal node in which it has landed. In \textit{regression trees}, this leaf value is a real number, and is the estimate of the response $y$ for the given covariates. Thus, regression trees are a nonparametric fitting procedure where the estimate of $f$ is a partition of predictor space into various hyperrectangles. Regression trees are well-known for their ability to approximate complicated response surfaces containing both nonlinearities and interaction effects.

There are many different ways to build decision trees. Many classic approaches rely on a greedy procedure to choose the best splitting rule at each node based on some pre-determined criterion. Once the construction of the tree is completed, the tree is then pruned back to prevent overfitting.

Previous efforts to handle missingness in trees include surrogate variable splitting \citep{Therneau1997}, \qu{Missing Incorporated in Attributes} \citep[MIA, ][section 2]{Twala2008} and many others (see \citealp{Ding2010} and \citealp{Twala2009}). MIA, the particular focus for this work, is a procedure that natively uses missingness when greedily constructing the rules for the decision tree's internal nodes. We summarize the procedure in Algorithm~\ref{alg:mia} and we explain how the expanded set of rules is injected into the \texttt{BART} procedure in Section~\ref{sec:mia_in_bart}.

\begin{algorithm}[htp]
\caption{\textit{Splitting rule choices during construction of a new tree branch in MIA.} \\ The algorithm chooses one of the following three rules for all splitting attributes and all splitting values $c$. Since there are $p$ splitting attributes and at most $n-1$ unique values to split on, the greedy splitting algorithm with MIA checks $2(n-1)p + p$ possible splitting rules at each iteration instead of the classic $(n-1)p$.}
\begin{algorithmic}[1]
\State If $x_{ij}$ is present and $x_{ij} \leq c$, send this observation left ($\longleftarrow$); otherwise, send this observation right ($\longrightarrow$). If $x_{ij}$ is missing, send this observation left ($\longleftarrow$).
\State If $x_{ij}$ is present and $x_{ij} \leq c$, send this observation left ($\longleftarrow$); otherwise, send this observation right ($\longrightarrow$). If $x_{ij}$ is missing, send this observation right ($\longrightarrow$).
\State If $x_{ij}$ is missing, send this observation left ($\longleftarrow$); if it is present, regardless of its value, send this observation right ($\longrightarrow$) .
\end{algorithmic}
\label{alg:mia}
\end{algorithm}

There are many advantages of the MIA approach. First, MIA has the ability to model complex MAR and NMAR relationships, as evidenced in both \cite{Twala2008} and the results of Sections~\ref{sec:generated_data_sims} and \ref{sec:real_data_sims}. Since missingness is integrated into the splitting rules, forecasts can be made without imputing when $\X_*$ contains missingness.

Another strong advantage of MIA is the ability to split on feature missingness (line 3 of Algorithm~\ref{alg:mia}). This splitting rule choice allows for the tree to better capture pattern mixture models where missingness directly influences the response model. Generally speaking, imputation ignores pattern mixture models; missingness is only viewed as holes to be filled-in and forgotten.

Due to these benefits as well as conceptual simplicity, we chose to implement MIA-within-\texttt{BART}, denoted \qu{\texttt{BARTm}}, when enhancing \texttt{BART} to handle missing data.

\subsection{BART}\label{subsec:bart}

Bayesian Additive Regression Trees is a combination of many regression trees estimated via a Bayesian model. Imagine the true response function can be approximated by the sum of $m$ trees with additive normal and homoskedastic noise:\footnote{\texttt{BART} can also be adapted for classification problems by using a probit link function and a data augmentation approach relying on latent variables \citep{Albert1993}.}

\bneqn\label{eq:bart}
\Y = f(\X) + \berrorrv \approx \treeleaft{1}(\X) + \treeleaft{2}(\X) + \ldots + \treeleaft{m}(\X) + \berrorrv, \quad\quad \berrorrv \sim \multnormnot{n}{\zerovec}{\sigsq\I_n}.
\eneqn

\noindent The notation, $\treeleaft{}$, denotes both structure and splitting rules ($\treet{}$) as well as leaf values ($\leaf$). 

\texttt{BART} can be distinguished from other purely algorithmic ensemble-of-trees models by its full Bayesian model, consisting of both a set of independent priors and likelihoods. Its posterior distribution is estimated via Gibbs Sampling \citep{Geman1984} with a Metropolis-Hastings step \citep{Hastings1970}.

There are three regularizing priors within the \texttt{BART} model which are designed to prevent overfitting. The first prior, placed on the tree structure is designed to prevent trees from growing too deep, thereby limiting the complexity that can be captured by a single tree. The second prior is placed on the leaf value parameters (the predicted values found in the terminal nodes) and is designed to shrink the leaf values towards the overall center of the response's empirical distribution. The third prior is placed on the variance of the noise $\sigsq$ and is designed to curtail overfitting by introducing noise into the model if it begins to fit too snugly. Our development of \texttt{BARTm} uses the default hyperparameters recommended in the original work \citep{Chipman2010}. For those who do not favor a pure Bayesian approach, these priors can be thought of as tuning parameters. 

In addition to the regularization priors, \texttt{BART} imposes an agnostic prior on the splitting rules within the decision tree branches. First, for a given branch, the splitting attribute is uniformly drawn from the set $\braces{x_1, \ldots, x_p}$ of variables available at the branch. The splitting value is then selected by drawing uniformly from the available values conditional on the splitting attribute $j$. Selecting attributes and values from a uniform discrete distribution represents a digression from the approach used in decision tree algorithms of greedily choosing splits based on some splitting criteria. Extending this prior allows for \texttt{BART} to incorporate MIA, which is discussed in Section \ref{sec:mia_in_bart}. 

To generate draws from the posterior distribution, each tree is fit iteratively, holding the other $m-1$ trees constant, by using only the portion of the response left unfitted. To sample trees, changes to the tree structure are proposed then accepted or rejected via a Metropolis-Hastings step. The tree proposals are equally-likely alterations: growing a leaf by adding two daughter nodes, pruning two twin leaves (rendering their parent node into a leaf), or changing a splitting rule. Following the tree sampling, the posterior for the leaf value parameters are Gibbs sampled. The likelihood of the predictions in each node is assumed to be normal. Therefore, the normal-normal conjugacy yields the canonical posterior normal distribution. After sampling all tree changes and terminal node parameters, the variance parameter $\sigsq$ is Gibbs sampled. By model assumption, the likelihood for the errors is normal and the conjugacy with the inverse-gamma prior yields the canonical posterior inverse-gamma.

Usually around 1,000 Metropolis-within-Gibbs iterations are run as ``burn-in'' until $\sigsq$ converges (by visual inspection). Another 1,000 or so are sampled to obtain \qu{burned-in} draws from the posterior, which define the \texttt{BART} model. Forecasts are then obtained by dropping the observations of $\X_*$ down the collection of sampled trees within each burned-in Gibbs sample. A point prediction $\hat{y}$ is generated by summing the posterior leaf values across the trees as in Equation \ref{eq:bart}. \textit{Credible intervals}, which are intervals containing a desired percentage (e.g. 95\%) of the posterior probability mass for a Bayesian parameter of interest, can be computed via the desired empirical quantiles over the burned-in samples.

For a thorough description about the internals of \texttt{BART} see \citet{Chipman2010} and \citet{Kapelner2013}.

\section{Missing Incorporated in Attributes within \texttt{BART}}\label{sec:mia_in_bart}

Implementing \texttt{BARTm} is straightforward. We previously described the prior on the splitting rules within the decision tree branches as being discrete uniform on the possible splitting attributes and discrete uniform on the possible splitting values. To account for Lines 1 and 2 in the MIA procedure (Algorithm \ref{alg:mia}), the splitting attribute $x_j$ and split value are proposed as explained in Section~\ref{subsec:bart}, but now we additionally propose a direction (left or right with equal probability) for records to be sent when the records have with missing values in $x_j$. A possible splitting rule would therefore be \qu{$x_{ij} <$ \texttt{c} and move left if $x_{ij}$ is missing.} To account for Line 3 in the algorithm, splitting on missingness itself, we create dummy vectors of length $n$ for each of the $p_M$ attributes with missingness, denoted $\M_1, \ldots, \M_{p_M}$, which assume the value 1 when the entry is missing and 0 when the entry is present. We then augment the original training matrix together with these dummies and use the augmented training matrix, $\Xtrain' := \bracks{\Xtrain, M_1, \ldots, M_{p_M}}$, as the training data in the \texttt{BARTm} algorithm. Once again, the prior on splitting rules is the same as the original \texttt{BART} but now with the additional consideration that the direction of missingness is equally likely left or right conditional on the splitting attribute and value. But why should this algorithm yield good predictive performance under the framework discussed in Section~\ref{subsec:overview}? 

We expect \texttt{BARTm} to exhibit greater predictive performance over MIA in classical decision trees for two reasons. First, \texttt{BARTm}'s sum-of-trees model offers much greater fitting flexibility of interactions and non-linearities compared to a single tree; this flexibility will explore models that the data analyst may not have thought of. Additionally, due to the greedy nature of decision trees, once a split is chosen, the direction in which missingness is sent cannot be reversed. \texttt{BARTm} can alter its trees by pruning and regrowing nodes or changing splitting rules. These proposed modifications to the trees are accepted or rejected stochastically using the Metropolis-Hastings machinery depending on how strongly the proposed move increases the model's likelihood. 

We hypothesize that \texttt{BARTm}'s stochastic search for splitting rules allows observations with missingness to be grouped with observations having similar response values. Due to the Metropolis-Hastings step, only splitting rules and corresponding groupings that increase overall model likelihood $\cprob{\Y}{\X, \M}$ will be accepted. In essence, \texttt{BARTm} is \qu{feeling around} predictor space for a location where the missing data would most increase the overall marginal likelihood. For selection models, since splitting rules can depend on any covariate including the covariate with missing data, it should be possible to generate successful groupings for the missing data under both MAR\footnote{As a simple example, suppose there are two covariates $X_1$ and $X_2$ and a MAR mechanism where $X_2$ is increasingly likely to go missing for large values of $X_1$. \texttt{BARTm} can partition this data in two steps to increase overall likelihood: (1) A split on a large value of $X_1$ and then (2) a split on $M_2$.} and NMAR\footnote{Suppose an NMAR mechanism where $X_2$ is more likely to be missing for large values of $X_2$. \texttt{BARTm} can select splits of the form \qu{$x_2 > c$ and $x_2$ is missing} with $c$ large. Here, the missing data is appropriately kept with larger values of $X_2$ and overall likelihood should be increased.} mechanisms. When missingness does not depend on any other covariates, it should be more difficult to find appropriate ways to partition the missing data,\footnote{Due to the regularization prior on the depths of the trees in \texttt{BART} coupled with the fact that all missing data must move to the same daughter node, the trees do not grow deep enough to create sufficiently complex partitioning schemes to handle the MCAR mechanism.} and we hypothesize that \texttt{BARTm} will be least effective for selection models with MCAR MDMs. Additionally, we hypothesize that \texttt{BARTm} has potential to perform well on pattern mixture models due to the partitioning nature of the regression tree. \texttt{BARTm} can partition the data based on different patterns of missingness by creating splits based on missingness itself. Then, underneath these splits, different submodels for the different patterns can be constructed. If missingness is related to the response, there is a good chance BARTm will find it and exploit it, yielding accurate forecasts.

Another motivation for adapting MIA to \texttt{BART} arises from computational concerns. \texttt{BART} is a computationally intensive algorithm, but its runtime increases negligibly in the number of covariates \citep[see][Section 6]{Chipman2010}. Hence, \texttt{BARTm} leaves the computational time virtually unchanged with the addition of the $p_M$ new missingness dummy covariates. Another possible strategy would be to develop an iterative imputation procedure using \texttt{BART} similar to that in \citet{Stekhoven2012} or a model averaging procedure using a multiple imputation framework, but we believe these approaches would be substantially more computationally intensive.

\section{Generated Data Simulations}\label{sec:generated_data_sims}

\subsection{A Simple Pattern Mixture Model}\label{subsec:pattern_mixture}

\begin{figure}[h]
\centering
\vspace{-0cm}
\begin{subfigure}[t]{0.65\textwidth}
\centering
\includegraphics[width=4.5in]{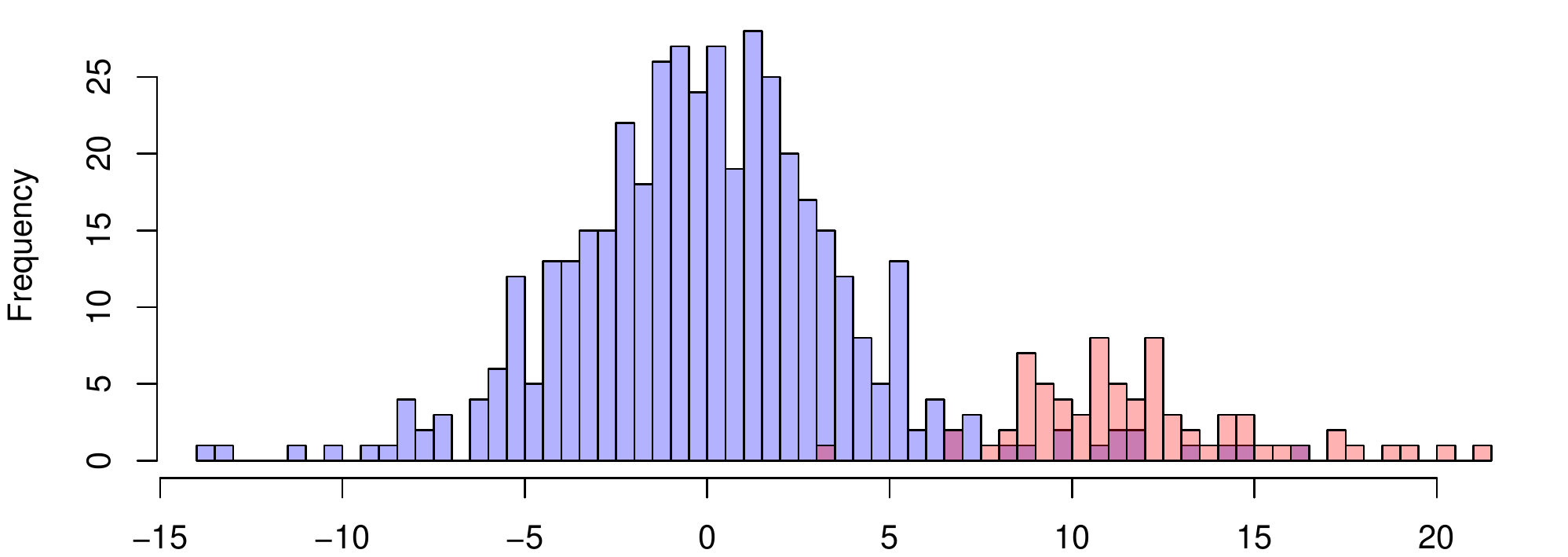}
\caption{\footnotesize Histogram of response values}
\label{subfig:model_hist_response}
\end{subfigure}~
\begin{subfigure}[t]{0.32\textwidth}
\centering
\includegraphics[width=2.0in]{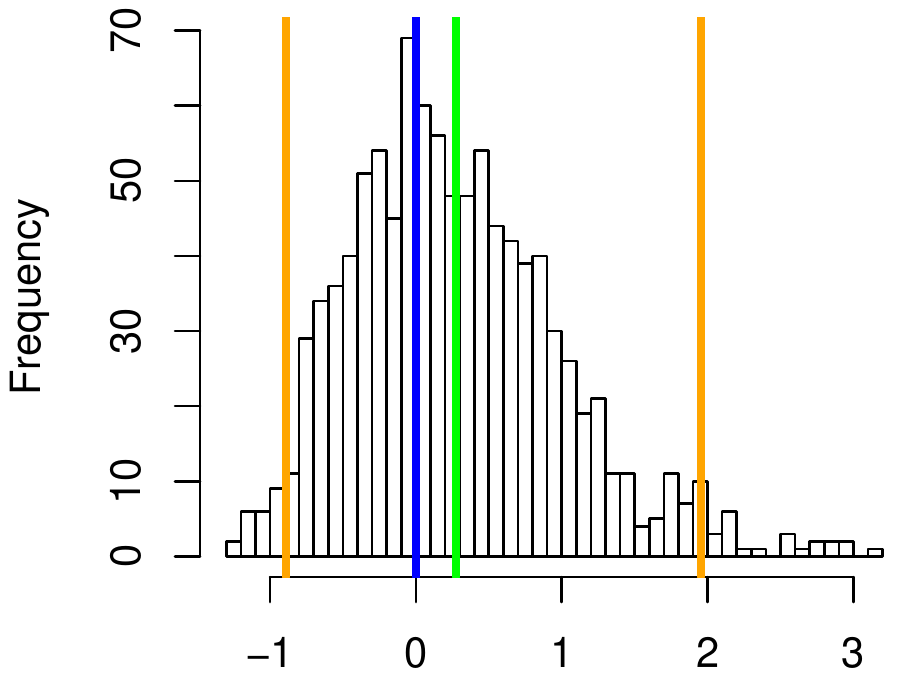}
\caption{\footnotesize $\x_* = \bracks{0~0~0},\hat{y} = 0.3 \pm 0.73$}
\label{fig:model000}
\end{subfigure}\\\vspace{0.2cm}
\begin{subfigure}[b]{0.33\textwidth}
\centering
\includegraphics[width=2.0in]{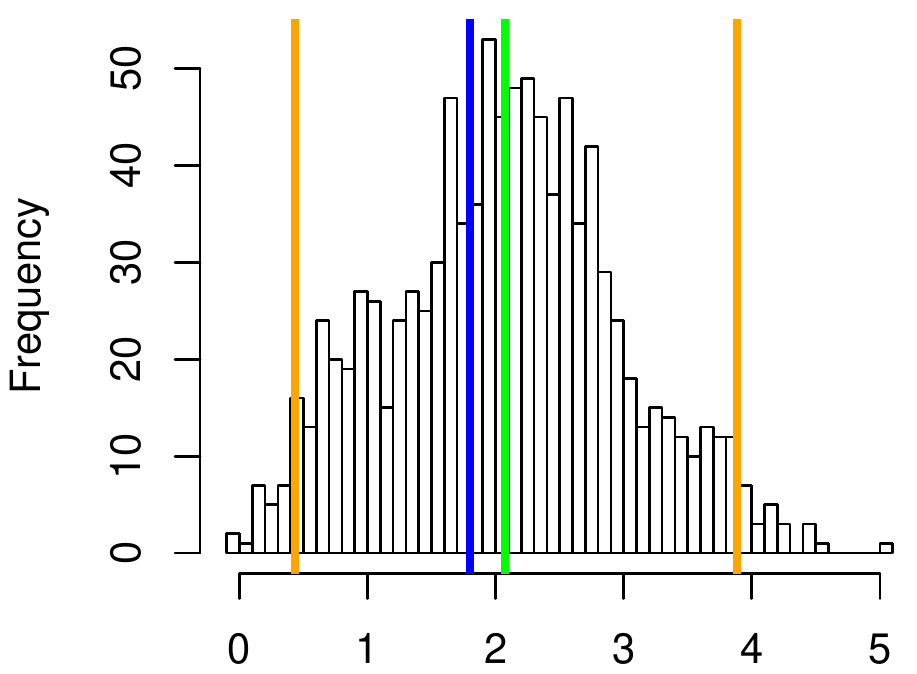}
\caption{\footnotesize $\x_* = \bracks{0~\cdot~0}, \hat{y} = 2.1 \pm 0.90$}
\label{fig:model0NA0}
\end{subfigure}{\tiny ~}
\begin{subfigure}[b]{0.33\textwidth}
\centering
\includegraphics[width=2.0in]{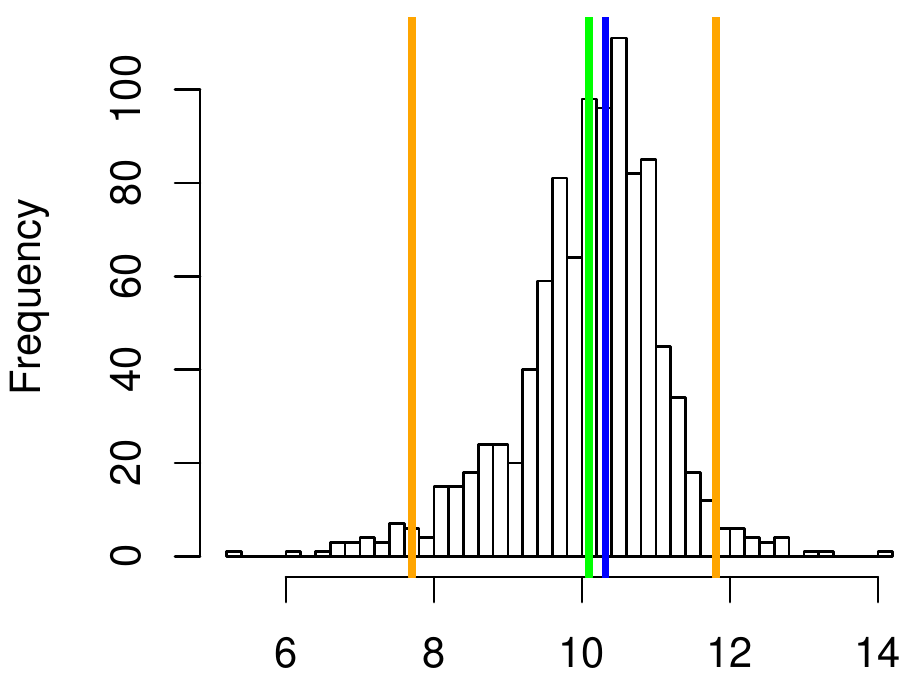}
\caption{\footnotesize $\x_* = \bracks{0~0~\cdot}, \hat{y} = 10.1 \pm 1.00$}
\label{fig:model00NA}
\end{subfigure}{\tiny ~}
\begin{subfigure}[b]{0.33\textwidth}
\centering
\includegraphics[width=2.0in]{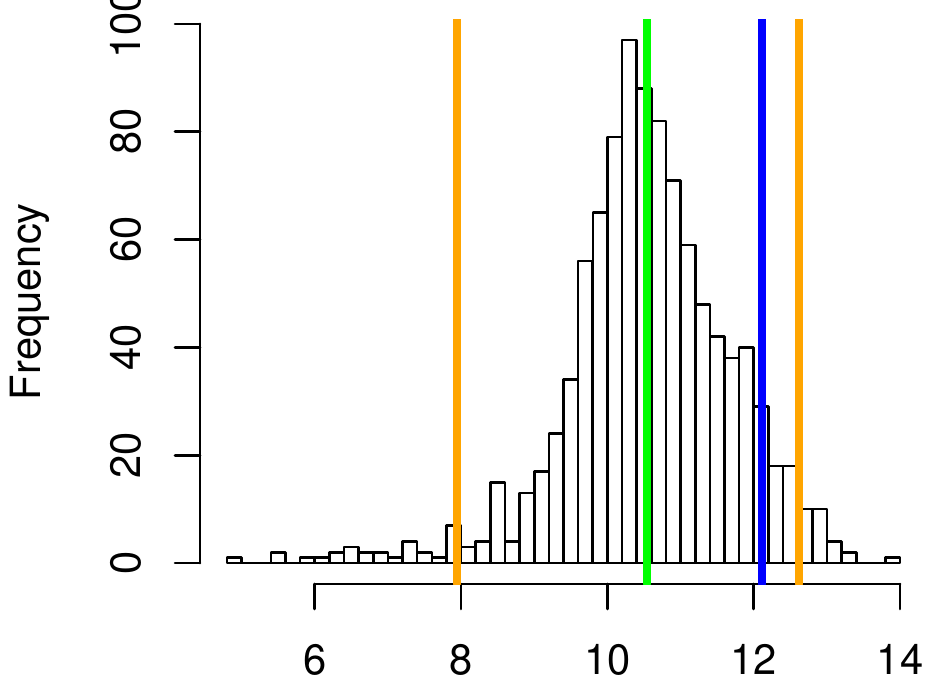}
\caption{\footnotesize $\x_* = \bracks{0~\cdot~\cdot}, \hat{y} = 10.5 \pm 1.14$}
\label{fig:model0NANA}
\end{subfigure}
\caption{(a) A $n = 500$ sample of the responses of the model in Equation~\ref{eq:pattern_mixture_model}. Colored in blue are the responses when $X_3$ is present and red are responses when $X_3$ is missing. (b-e) 1,000 burned-in posterior draws from a \texttt{BARTm} model for different values of $\x_*$ drawn from the data generating process found in Equation~\ref{eq:pattern_mixture_model}. The \ingreen{green} line is \texttt{BARTm}'s forecast $\hat{y}$ (the average of the posterior burned-in samples). The \inblue{blue} line is the true model expectation. The two \inyellow{yellow} lines are the bounds of the 95\% credible interval for $\cexpe{\Y}{\Xobs=\x_*,~\M=\bv{m}_*}$.}
\label{fig:pattern_mixture_pred_and_ci_results}
\end{figure}

We begin with an illustration of \texttt{BARTm}'s ability to directly estimate $\cexpe{\Y}{\Xobs, \M}$ and additionally provide uncertainty intervals. We consider the following nonlinear response surface:

\bneqn\label{eq:pattern_mixture_model}
&& \Y = g(X_1, X_2, X_3) + \B M_3 + \berrorrv, \quad \errorrv \iid \normnot{0}{\sigsq_e} \quad B \iid \normnot{\mu_b}{\sigsq_b}\\
&& g(X_1, X_2, X_3) = X_1 + X_2 + 2 X_3 - X_1^2 + X_2^2 + X_1 X_2 \nonumber \\ 
&& \bracks{X_1, X_2, X_3} \iid \multnormnot{3}{\zerovec}{\sigsq_x \threebythreemat{1}{\rho_1}{\rho_2}{\rho_1}{1}{\rho_1}{\rho_2}{\rho_1}{1}}, \nonumber 
\eneqn

\noindent where $\sigsq_x = 1,~\rho_1 = 0.2,~\rho_2 = 0.4,~\sigsq_e = 1,~\mu_b = 10$ and $\sigsq_b = 0.5$. Note that the pattern mixture model is induced by missingness in $X_3$. Under this missingness pattern, the response is offset by $B$, a draw from a normal distribution. Figure~\ref{subfig:model_hist_response} displays the $n = 500$ sample of the response from the model colored by $M_3$ to illustrate the separation of the two response patterns. We choose the following jointly NMAR MDM for $X_2$ and $X_3$ which were chosen to be simple for the sake of ensuring that the illustration is clear. The next section features more realistic mechanisms.

\bneqn\label{eq:pattern_mixture_model_mdms}
&& 1: X_2 ~~\text{is missing with probability 0.3 if}~~ X_2 \geq 0 \\
&& 2: X_3 ~~\text{is missing with probability 0.3 if}~~ X_1 \geq 0. \nonumber
\eneqn

If the \texttt{BARTm} model assumptions hold and is successfully able to estimate $\cexpe{\Y}{\Xobs, \M}$, then the true $\cexpe{\Y}{\Xobs, \M}$ is highly likely to be contained within a 95\% credible interval for the prediction. We first check to see whether \texttt{BARTm} can capture the correct response when $\Xtrain$ has missing entries but $\X_*$ does not. Predicting on $\x_* = \bracks{0~0~0}$ should give $\cexpe{\Y}{\X = \x_*} = 0$ for the prediction. Figure~\ref{fig:model000} illustrates that \texttt{BARTm} captures the expected value within its 95\% credible interval.

Next we explore how well \texttt{BARTm} estimates the conditional expectation when missingness occurs within the new observation $\x_*$. We examine how \texttt{BARTm} handles missingness in attribute $X_2$ by predicting on $\x_* = \bracks{0~\cdot~0}$ where the \qu{$\cdot$} denotes missingness. By Equation~\ref{eq:pattern_mixture_model_mdms}, $X_2$ is missing 30\% of the time if $X_2$ itself is greater than 0. By evaluating the moments of a truncated normal distribution, it follows that \texttt{BARTm} should guess $\cexpe{X_2 + X_2^2}{X_2 > 0} = \sqrt{2 / \pi} + 1 \approx 1.80$. Figure~\ref{fig:model0NA0} indicates that \texttt{BARTm}'s credible interval captures this expected value. Note the larger variance of the posterior distribution relative to Figure~\ref{fig:model000} reflecting the higher uncertainty due to $x_{*_2}$ going missing. This larger interval is a great benefit of MIA. As the trees are built during the Gibbs sampling, the splitting rules on $X_2$ are accompanied by a protocol for missingness: missing data will flow left or right in the direction that increases model likelihood and this direction is chosen with randomness. Thus, when $\x_*$ is predicted with $x_{*_2}$ missing, missing records flow left and right over the many burned-in Gibbs samples creating a wider distribution of predicted values, and thus a wider credible interval. This is an important point --- \texttt{BARTm} can give a rough estimate of how much information is lost when values in new records become missing by looking at the change in the standard error of a predicted value.\footnote{Note that if \texttt{BART}'s hyperparameters are considered \qu{tuning parameters,} the credible intervals endpoints are not directly interpretable. However, the relative lengths of the intervals can be trusted whereby they signify different levels of forecast confidence to the practitioner.}

We next consider how \texttt{BARTm} performs when $X_3$ is missing by predicting on $\x_* = \bracks{0~0~\cdot}$. By Equation \ref{eq:pattern_mixture_model_mdms}, \texttt{BARTm} should guess $\cexpe{X_3}{X_1 > 0} = .4\sqrt{2 / \pi} \approx .32$ (which follows directly from the properties of the conditional distribution of bivariate normal distribution, recalling that $\corr{X_1}{X_3} = 0.4$). When $X_3$ is missing, there is a different response pattern, and the response is shifted up by $B$. Since $\expe{B} = 10$, \texttt{BARTm} should predict approximately 10.32. The credible interval found in Figure \ref{fig:model00NA} indicates that \texttt{BARTm}'s credible interval again covers the conditional expectation.

Finally, we consider the case where $X_2$ and $X_3$ are simultaneously missing. Predicting on $\x_* = \bracks{0~\cdot~\cdot}$ has a conditional expectation of $\cexpe{X_2 + X_2^2}{X_2 > 0} + \cexpe{X_3}{X_1 > 0} + \expe{B} \approx 12.12$. Once again, the posterior draws displayed in Figure \ref{fig:model0NANA} indicate that \texttt{BARTm} reasonably estimates the conditional expectation. Note that the credible interval here is wider than in Figure \ref{fig:model00NA} due to the additional missingness of $X_2$.

\subsection{Selection Model Performance}\label{subsec:selection_model_performance}

In order to gauge \texttt{BARTm}'s out-of-sample predictive performance on selection models and to evaluate the improvement over model-building on complete cases, we construct the same model as Equation~\ref{eq:pattern_mixture_model} withholding the offset $B$ (which previously induced the pattern mixture). Thus $\Y = g(X_1, X_2, X_3) + \berrorrv$. We imposed three scenarios illustrating performance under the following missingness mechanisms. The first is MCAR; $X_1$ is missing with probability $\gamma$. The second is MAR; $X_3$ is missing according to a non-linear probit model depending on the other two covariates:

\bneqn\label{eq:crazy_model_mar}
\cprob{M_{3} = 1}{X_1, X_2} = \Phi\parens{\gamma_0 + \gamma_1 X_{1} + \gamma_1 X_{2}^2}.
\eneqn

\noindent The last is NMAR; $X_2$ goes missing according to a similar non-linear probit model this time depending on itself and $X_1$:

\bneqn\label{eq:crazy_model_mar}
\cprob{M_{2} = 1}{X_1, X_2} = \Phi\parens{\gamma_0 + \gamma_1 X_{1}^2 + \gamma_1 X_{2}}.
\eneqn

For each simulation, we set the number of training observations to $n = 250$ and simulate 500 times. Additionally, each simulation is carried out with different levels of missing data, approximately $\braces{0, 10, \ldots, 70}$ percent of rows have at least one missing covariate entry. For the MCAR dataset, the corresponding $\gamma = \braces{0, 0.03,0.07, 0.11, 0.16, 0.26, 0.33}$ and for both the MAR and NMAR datasets it was $\gamma_0 = -3$ and $\gamma_1 = \braces{0, 0.8, 1.4, 2.0, 2.7, 4.0, 7.0, 30}$.

We record results for four different scenarios: (1) $\Xtrain$ and $\X_*$ contain missingness (2) $\Xtrain$ contains missingness and $\X_*$ is devoid of missing data\footnote{In this case, $\X_*$ is generated without the MDM to maintain a constant number of rows.} (3) only complete cases of $\Xtrain$ are used to build the model but $\X_*$ contains missingness and (4) only complete cases of $\Xtrain$ are used to build the model and $\X_*$ is devoid of missing data. 

\begin{figure}[htp]
\centering
\begin{subfigure}[b]{0.32\textwidth}
\centering
\includegraphics[width=2.1in]{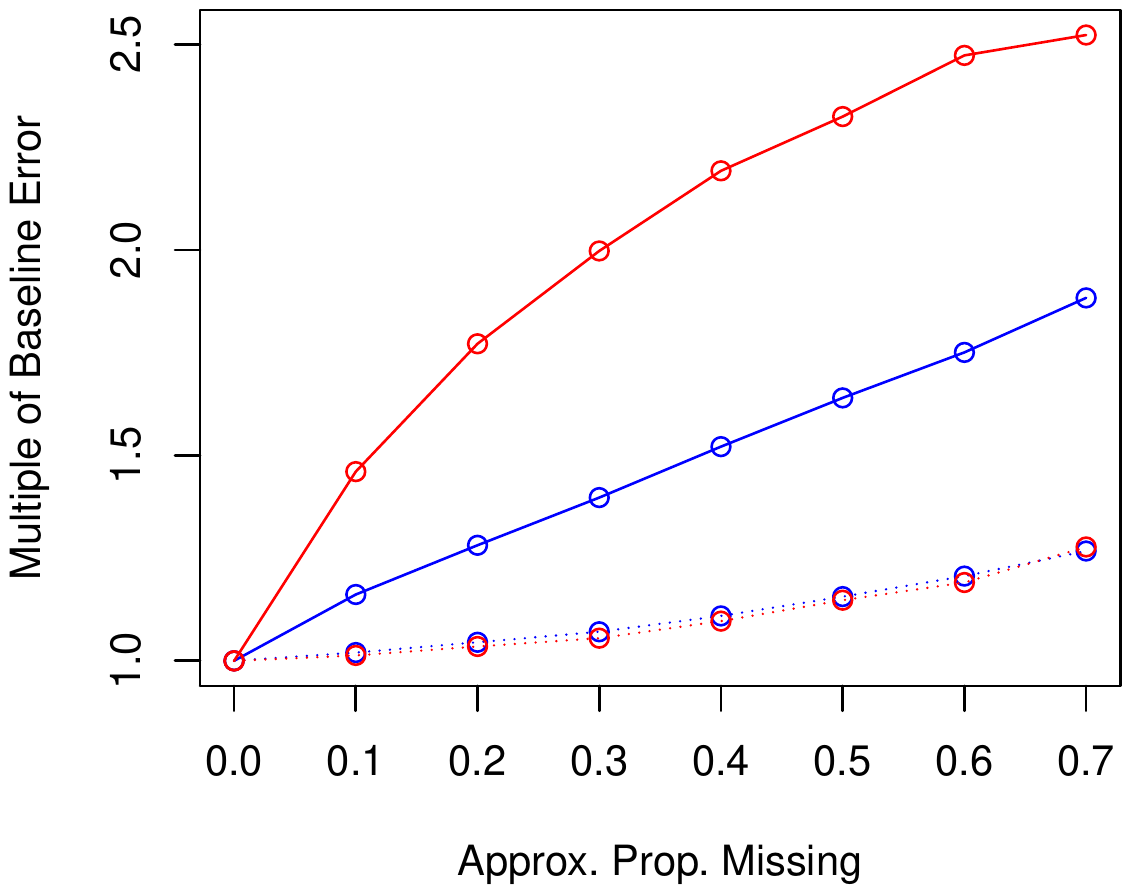}
\caption{MCAR}
\label{subfig:selection_model_sims_mcar}
\end{subfigure}~
\begin{subfigure}[b]{0.32\textwidth}
\centering
\includegraphics[width=2.0in]{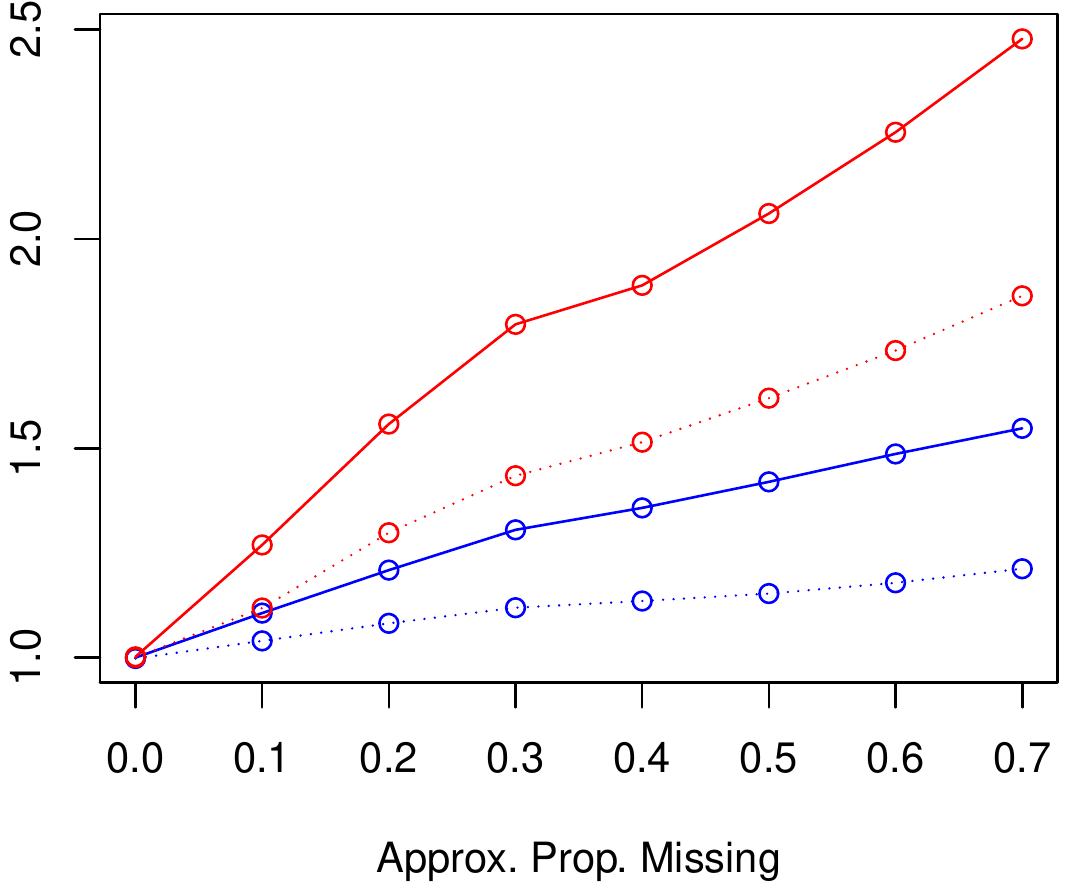}
\caption{MAR}
\label{subfig:selection_model_sims_mar}
\end{subfigure}~
\begin{subfigure}[b]{0.32\textwidth}
\centering
\includegraphics[width=2.0in]{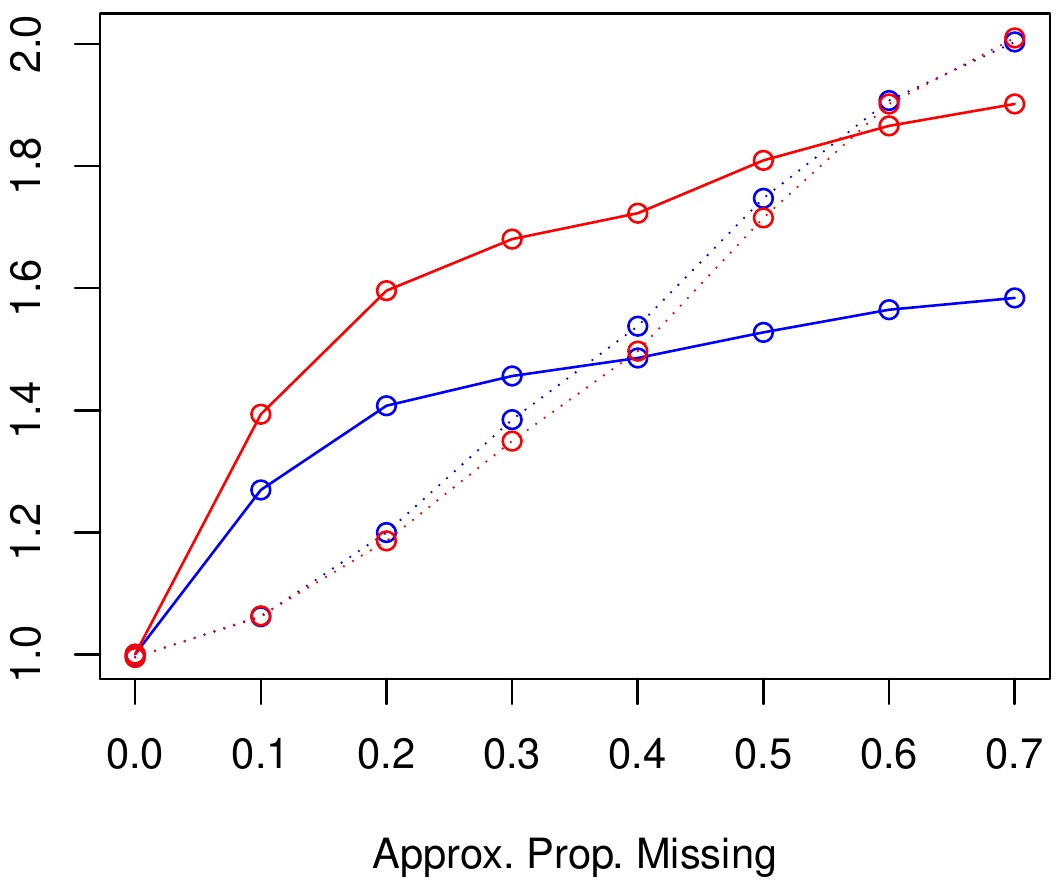}
\caption{NMAR}
\label{subfig:selection_model_sims_nmar}
\end{subfigure}
\caption{Simulation results of the response model for the three MDM's explained in the text. The $y$-axis measures the multiple of out-of-sample root mean square error (oosRMSE) relative to the performance under the absence of missingness. \inblue{Blue} lines correspond to the two scenarios where \texttt{BART} was built with all cases in $\Xtrain$ and \inred{red} lines correspond to the two scenarios where \texttt{BART} was built with only the complete cases of $\Xtrain$. Solid lines correspond to the two scenarios where $\X_*$ included missing data and dotted lines correspond to the two scenarios where the MDM was turned off in $\X_*$.}
\label{fig:selection_model_sims}
\end{figure}

We make a number of hypotheses about the relationship between the predictive performance of using incomplete cases (all observations) compared to the complete case performance. As we discussed in Section~\ref{sec:mia_in_bart}, \texttt{BARTm} should be able model the expectation of the marginal likelihood in selection models, thus we expect models built with incomplete cases to predict better than models that were built with only the complete cases. The complete case models suffer from performance degradation for two main reasons (1) these models are built with a smaller sample size and hence their estimate of $\cexpe{\Y}{\Xobs, \M}$ is higher in bias and variance (2) the lack of missingness during the training phase does not allow the model to learn how to properly model the missingness, resulting in the missing data being filtered randomly from node to node during forecasting. These hypotheses are explored in Figure \ref{fig:selection_model_sims} by comparing the solid blue and solid red lines.

Further, during forecasting, we expect $\X_*$ samples with incomplete cases to have worse performance than the full $\X_*$ samples (devoid of missingness) simply because missingness is equivalent to information loss. However, for the NMAR model, we expect prediction performance on $\X_*$ without missingness to eventually, as the amount of missingness increases, be beaten by the predictive performance on $\X_*$ with missingness. Eventually there will be so much missingness in $X_2$ that (1) the trained model on missingness will only be able to create models by using $\M_2$ and expect $\M_2$ in the future $\X_*$ and (2) the trained model on complete cases will never observe the response of the function where $X_2$ went missing. These hypotheses are explored in Figure~\ref{fig:selection_model_sims} by comparing the solid lines to the dashed lines within the same color.

The results for the four scenarios under the three MDM's comport with our hypotheses. The solid red line is uniformly higher than the solid blue line confirming degradation for complete-case model forecasting on data with missingness. The dotted lines are lower than their solid counterparts indicating that providing more covariate information yields higher predictive accuracy. The one exception is for NMAR. After the number of rows with missingness is more than 40\%, forecasts on complete-cases only begin to perform worse than the forecast data with missingness for models built with missingness (\texttt{BARTm}).

In conclusion, for this set of simulations, \texttt{BARTm} performs better than \texttt{BART} models that ignore missingness in the training phase. The next section demonstrates \texttt{BARTm}'s performance in a real data set and compares its performance to a non-parametric statistical learning procedure that relies on imputation.

\section{Real Data Example}\label{sec:real_data_sims}

The Boston Housing data (BHD) measures 14 features about housing in the $n=506$ census tracts in Boston in 1970. For model building, the response variable is usually taken to be the median home value. For this set of simulations, we evaluate the performance of three procedures (1) \texttt{BARTm} (2) \texttt{RF} with $\Xtrain$ and $\X_*$ imputed via \texttt{MissForest} and (3) \texttt{BART} with $\Xtrain$ and $\X_*$ imputed via \texttt{MissForest}.\footnote{We assume \textit{a priori} that $\X_*$ will have missing data. Thus, the complete-case comparisons a la Section~\ref{subsec:selection_model_performance} were not possible.} We gauge out-of-sample predictive performance as measured by the oosRMSE for the three procedures on the simulation scenarios described in Table \ref{tab:bhd_scenarios}.

\begin{table}[htp]
\centering
\begin{tabular}{l|l}
Scenario & Description \\ \hline
Selection Model MCAR & \texttt{rm}, \texttt{crim}, \texttt{lstat}, \texttt{nox} and \texttt{tax} are each missing w.p. $\gamma$ \\
Selection Model MAR & \texttt{rm} and \texttt{crim} are missing according to the following models: \\
& $\prob{\M_{\text{\texttt{rm}}} = 1} = \Phi(\gamma_0 + \gamma_1 (\texttt{indus} + \texttt{lstat} + \texttt{age}))$ \\
& $\prob{\M_{\text{\texttt{crim}}} = 1} = \Phi(\gamma_0 + \gamma_1 (\texttt{nox} + \texttt{rad} + \texttt{tax}))$ \\
Selection Model NMAR & \texttt{rm} and \texttt{crim} are missing according to the following models: \\
& $\prob{\M_{\text{\texttt{rm}}} = 1} = \Phi(\gamma_0 + \gamma_1 (\texttt{rm} + \texttt{lstat}))$ \\
& $\prob{\M_{\text{\texttt{crim}}} = 1} = \Phi(\gamma_0 + \gamma_1 (\texttt{crim} + \texttt{nox}))$ \\
Pattern Mixture & The MAR selection model above and two offsets: \\
& (1) if $\M_{\text{\texttt{rm}}} = 1$, the response is increased by $\normnot{\mu_b}{\sigsq_b}$ \\
& (2) if $\M_{\text{\texttt{crim}}} = 1$, the response is decreased by $\normnot{\mu_b}{\sigsq_b}$
\end{tabular}
\caption{Missingness scenarios for the BHD simulations. Monospace \texttt{codes} are names of covariates in the BHD. Note that \texttt{rm} has sample correlations with \texttt{indus}, \texttt{lstat} and \texttt{age} of -0.39, -0.61 and -0.24 and \texttt{crim} has sample correlations with \texttt{nox}, \texttt{rad}, and \texttt{tax} of 0.42, 0.63 and 0.58. These high correlations should allow for imputations that perform well.}
\label{tab:bhd_scenarios}
\end{table}

Similar to Section~\ref{subsec:selection_model_performance}, each simulation is carried out with different levels of missing data, approximately $\braces{0, 10, 20, \ldots, 70}$ percent of rows have at least one missing covariate entry. For the MCAR scenario, the corresponding $\gamma = \braces{0, 0.02, 0.04, 0.07, 0.10, 0.13, 0.17}$, for the MAR scenario and pattern mixture scenario, $\gamma_1 = \braces{0, 1.3, 1.5, 1.7, 2.1, 2.6, 3.1, 3.8}$ and $\gamma_0$ is constant at -3 and for the NMAR scenario $\gamma_1 = \braces{0, 3.3, 3.6, 3.9, 4.1, 4.3, 4.6, 4.8}$ and $\gamma_0$ is constant at -3. Similar to Section~\ref{subsec:pattern_mixture}, we induce a pattern mixture model by creating a normally distributed offset based on missingness (we create two such offsets here). Here, we choose $\mu_b$ to be 25\% of the range in $y$ and $\sigma_b$ to be $\mu_b / 4$. These values are arbitrarily set for illustration purposes. It is important to note that the performance gap of \texttt{BARTm} versus \texttt{RF} with imputation can be arbitrarily increased by making $\mu_b$ larger.

For each scenario and each level of missing data, we run 500 simulations. In each simulation, we first draw missingness via the designated scenario found in Table~\ref{tab:bhd_scenarios}. Then, we randomly partition 80\% of the 506 BHD observations (now with missingness) as $\Xtrain$ and the remaining 20\% as $\X_*$. We build all three models (\texttt{BARTm}, \texttt{RF} with \texttt{MissForest} and \texttt{BART} with \texttt{MissForest}) on $\Xtrain$, forecast on $\X_*$ and record the oosRMSE. Thus, we integrate over idiosyncrasies that could be found in a single draw from the MDM and idiosyncrasies that could be found in a single train-test partition. When using \texttt{MissForest} during training, we impute values for the missing entries in $\Xtrain$ using $\bracks{\Xtrain, \ytrain}$ column-binded together. To obtain forecasts, we impute the missing values in $\X_*$ using $\bracks{\Xtrain, \X_*}$ row-binded together then predict using the bottom rows (i.e. those corresponding to the imputed test data). Note that we use \texttt{MissForest} in both \texttt{RF} and \texttt{BART} to ensure that the difference in predictive capabilities of \texttt{BART} and \texttt{RF} are not driving the results. 

For the MCAR selection model, we hypothesize that the \texttt{MissForest}-based imputation procedures will outperform \texttt{BARTm} due to the conceptual reasons discussed in Section~\ref{sec:mia_in_bart}. For the MAR selection model, we hypothesize similar performance between \texttt{BARTm} and both \texttt{MissForest}-based imputation procedures, as both MIA and imputation are designed to perform well in this scenario. In the NMAR selection model and pattern mixture model, we hypothesize that \texttt{BARTm} will outperform both \texttt{MissForest}-based imputation procedures, as \texttt{MissForest} (1) cannot make use of the values in the missingness columns it is trying to impute and (2) cannot construct different submodels based on missingness. Although imputation methods are not designed to handle these scenarios, it is important to run this simulation to ensure that \texttt{BARTm}, which \textit{is} designed to succeed in these scenarios, has superior out-of-sample predictive performance.

The results displayed in Figure \ref{fig:bhd_sims} largely comport with our hypotheses. \texttt{MissForest}-based methods perform better on the MCAR selection model scenario (Figure \ref{fig:mcar_bhd}) and \texttt{BARTm} is stronger in the NMAR scenario (Figure \ref{fig:nmar_bhd}) and pattern mixture scenario (Figure \ref{fig:bump_bhd}). It is worth noting that in the MAR selection model scenario (Figure \ref{fig:mar_bhd}), \texttt{BARTm} begins to outperform the imputation-based methods once the percentage of missing data becomes greater than 20\%. The performance of the imputation-based algorithms degrades rapidly here, while \texttt{BARTm}'s performance remains fairly stable, even with 70\% of the rows having at least one missing entry. In conclusion, \texttt{BARTm} will generally perform better than \texttt{MissForest} because it is not \qu{limited} to what can be imputed from the data on-hand. This advantage generally grows with the amount of missingness.

\newcommand{\bhdfigwidth}{3.1in}
\begin{figure}[htp]
\centering
\begin{subfigure}[b]{0.49\textwidth}
\centering
\includegraphics[width=\bhdfigwidth]{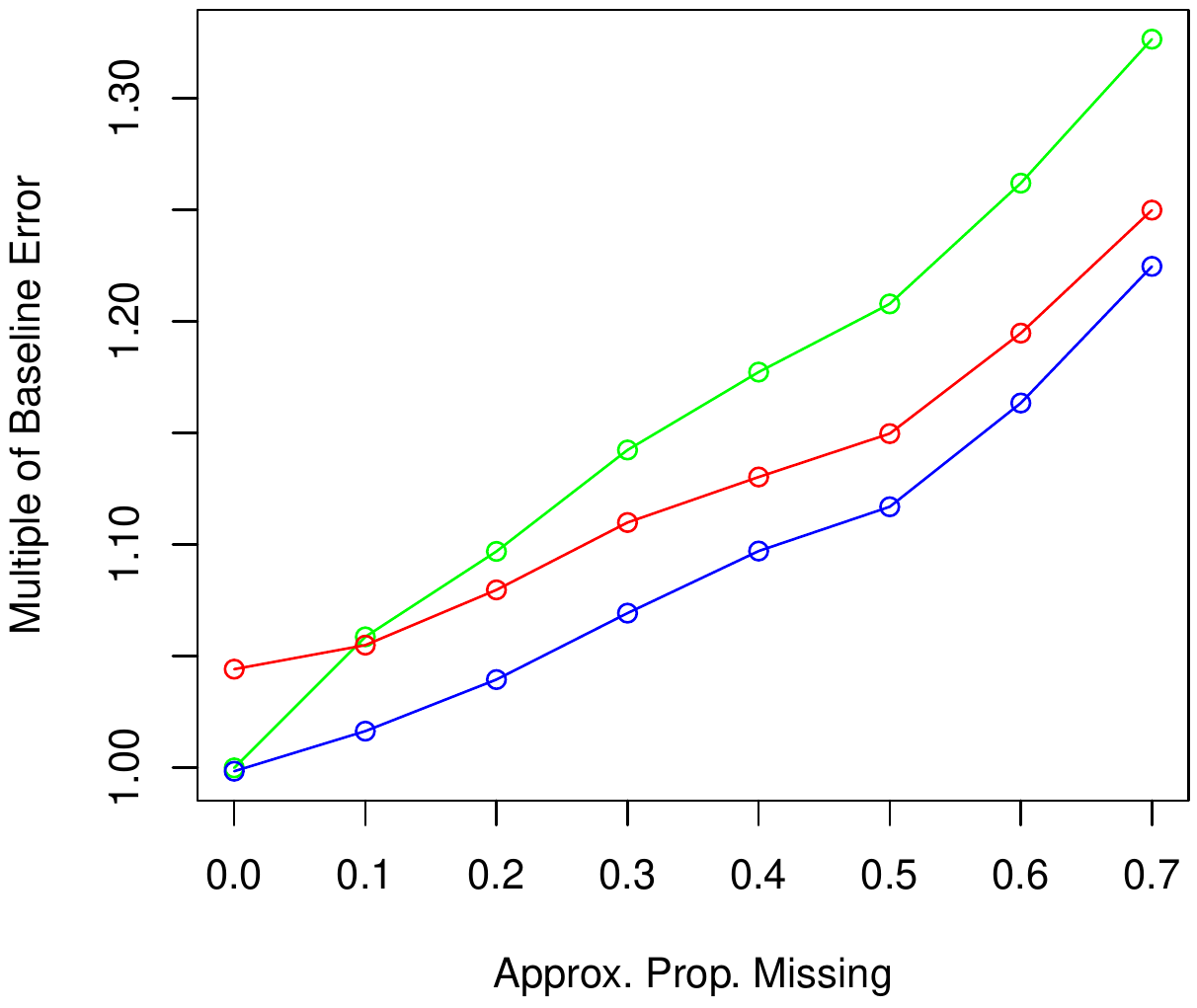}
\caption{MCAR}
\label{fig:mcar_bhd}
\end{subfigure}~~~
\begin{subfigure}[b]{0.49\textwidth}
\centering
\includegraphics[width=\bhdfigwidth]{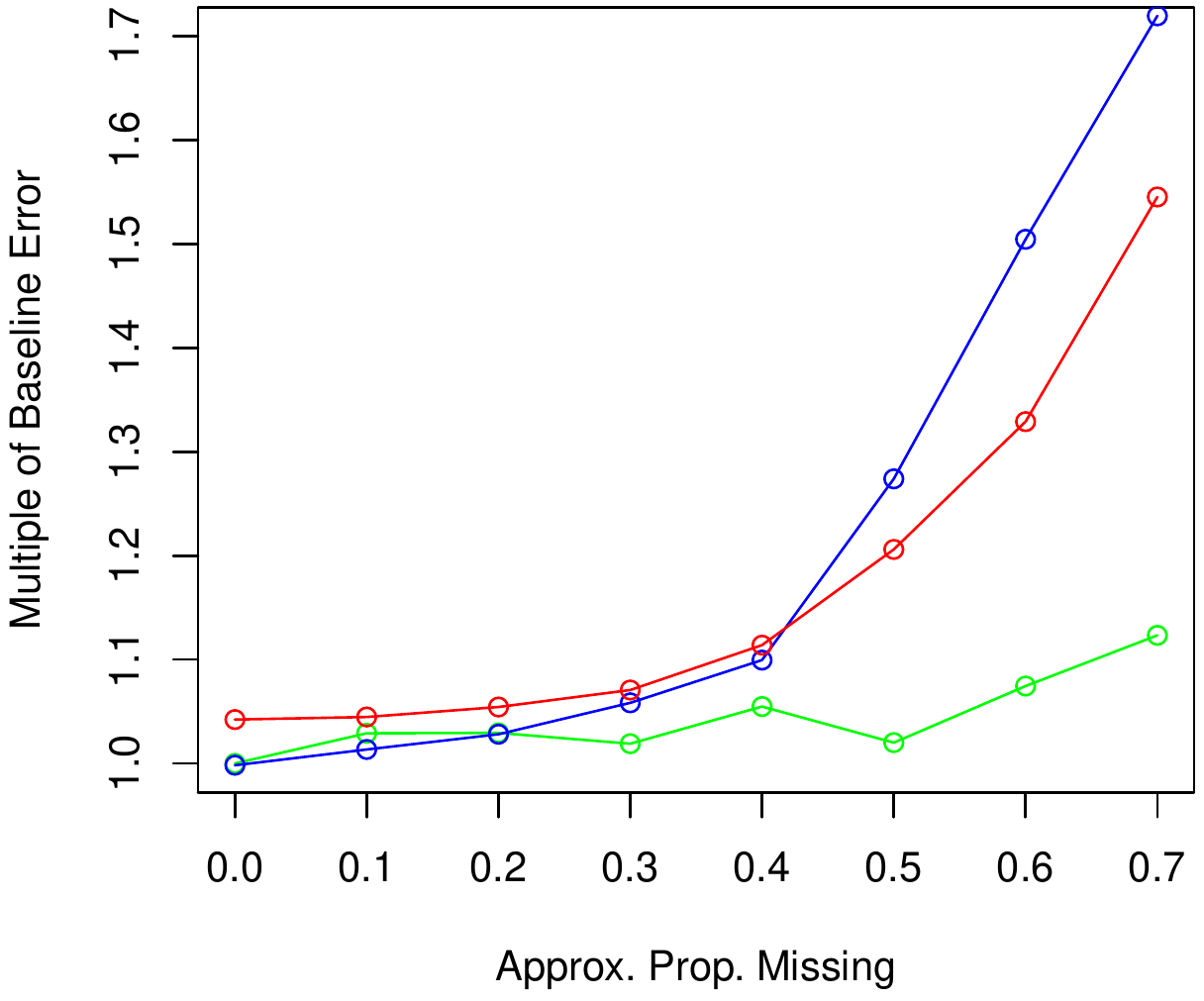}
\caption{MAR}
\label{fig:mar_bhd}
\end{subfigure}\\
\begin{subfigure}[b]{0.49\textwidth}
\centering
\includegraphics[width=\bhdfigwidth]{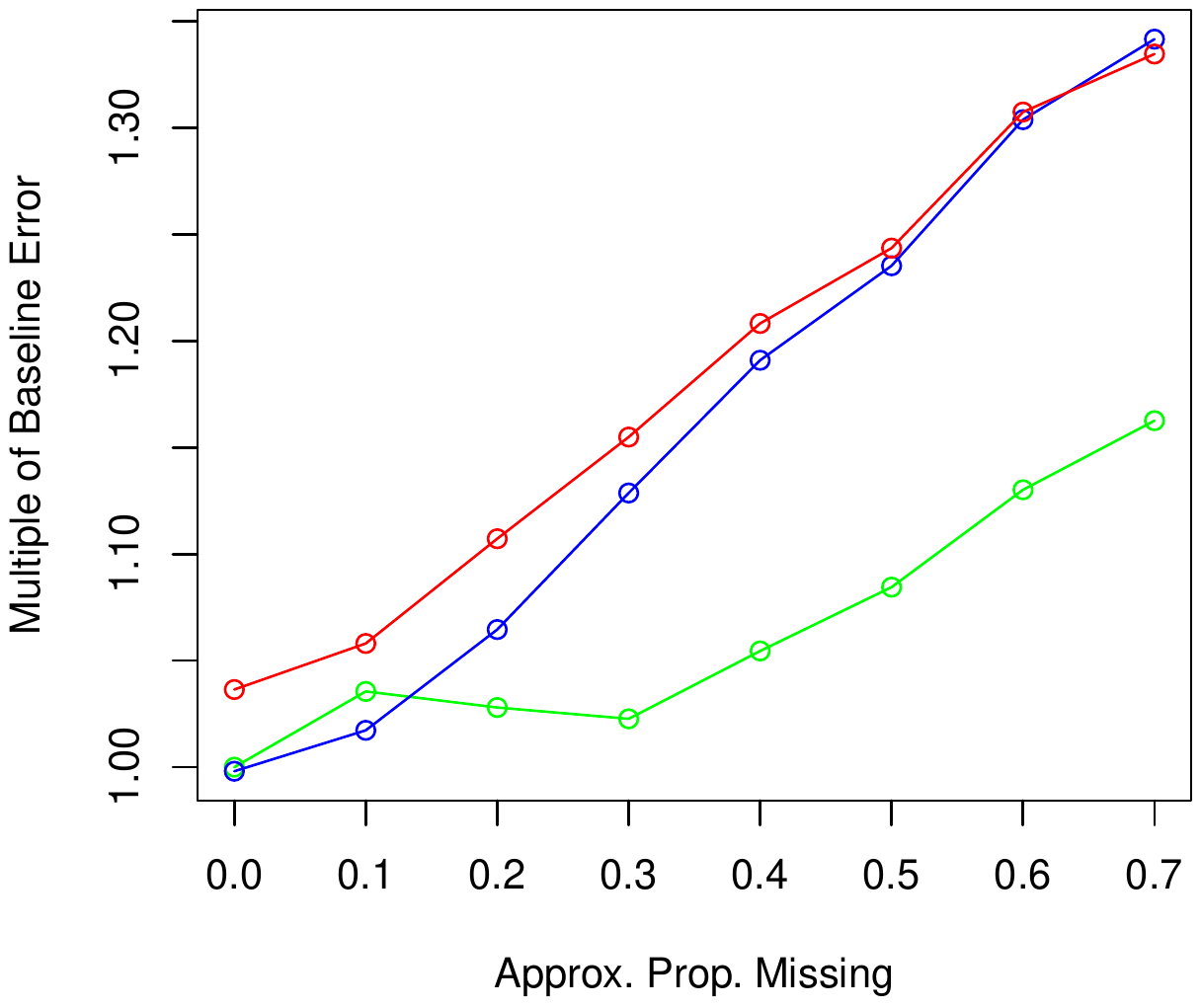}
\caption{NMAR}
\label{fig:nmar_bhd}
\end{subfigure}~~~
\begin{subfigure}[b]{0.49\textwidth}
\centering
\includegraphics[width=\bhdfigwidth]{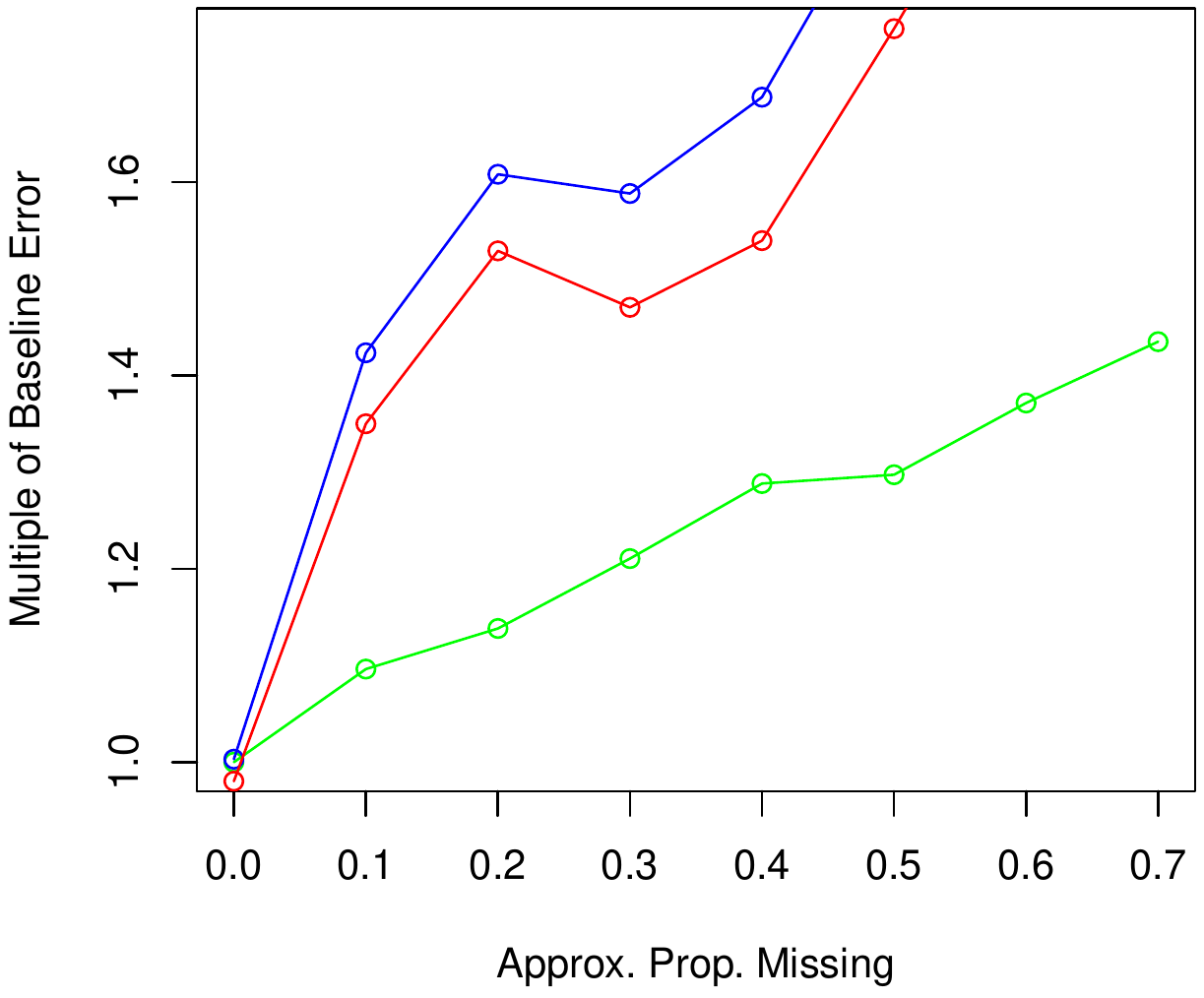}
\caption{Pattern Mixture}
\label{fig:bump_bhd}
\end{subfigure}
\caption{Simulations for different probabilities of missingness across the four simulated missing data scenarios in the BHD. The y-axis is oosRMSE relative to \texttt{BART}'s oosRMSE on the full dataset. Lines in \ingreen{green} plot \texttt{BARTm}'s performance, lines in \inred{red} plot \texttt{RF}-with-\texttt{MissForest}'s performance, and lines in \inblue{blue} plot \texttt{BART}-with-\texttt{MissForest}'s performance. Note that the \texttt{MissForest}-based imputation might perform worse in practice because here we allow imputation of the entire test set. In practice, it is likely that test observations appear sequentially.}
\label{fig:bhd_sims}
\end{figure}

\section{Discussion}\label{sec:discussion}

We propose a means of incorporating missing data into statistical learning for prediction problems where missingness may appear during both the training and forecasting phases. Our procedure, \texttt{BARTm}, implements \qu{missing incorporated in attributes} (MIA), a technique recently explored for use in decision trees, into Bayesian Additive Regression Trees, a newly developed tree-based statistical learning algorithm for classification and regression. MIA natively incorporates missingness by sending missing observations to one of the two daughter nodes. Due to the Bayesian framework and the Metropolis-Hastings sampling, missingness is incorporated into splitting rules which are chosen to increase overall model likelihood. This innovation allows missingness itself to be used as a legitimate value within splitting criteria, resulting in no need for imputing in the training or new data and no need to drop incomplete cases.

For the simulations explored in this article, \texttt{BARTm}'s performance was superior to models built using complete cases, especially when missingness appeared in the test data as well. Additionally, \texttt{BARTm} provided higher predictive performance on the MAR selection model relative to \texttt{MissForest}, a non-parametric imputation technique. We also observe promising performance on NMAR selection models and pattern mixture models in simulations. To the best of our knowledge, there is no clear evidence of other techniques that will exhibit uniformly better predictive performance in both selection and pattern mixture missingness models. Additionally, \texttt{BARTm}'s Bayesian nature provides informative credible intervals reflecting uncertainty when the forecasting data has missing covariates. 

Although the exploration in this article was focused on regression, we have observed \texttt{BARTm} performing well in binary classification settings. \texttt{BARTm} for both classification and regression is implemented in the \texttt{R} package \texttt{bartMachine}.

Due to MIA's observed promise, we recommend it as a viable strategy to handle missingness in other tree-based statistical learning methods. Future work should also consider exploration of methods that combine imputation with MIA appropriately, in order to enhance predictive performance for MCAR missing data mechanisms.

\section*{Supplementary Materials}

\begin{description}
\item[Computer Code:] Simulated results, tables, and figures can be replicated via the scripts located at \url{http://github.com/kapelner/bartMachine} (the home of the CRAN package \texttt{bartMachine}) in the \texttt{missing\_data\_paper} folder.
\end{description}

\section*{Acknowledgements}

We thank Richard Berk, Dana Chandler, Ed George, Dan Heitjan, Shane Jensen, and Jos{\'e} Zubizaretta for helpful discussions. We thank the anonymous reviewers for their helpful comments. Adam Kapelner acknowledges support from the National Science Foundation's Graduate Research Fellowship Program.

\bibliographystyle{plainnat}
\bibliography{bartm_paper}

\end{document}